\documentclass[10pt,twocolumn,letterpaper]{article}

\usepackage[pagenumbers]{cvpr} 
\usepackage[pagebackref,breaklinks,colorlinks]{hyperref}
\usepackage{graphicx}
\usepackage{amsmath}
\usepackage{amssymb}
\usepackage{booktabs}
\usepackage{multirow}
\usepackage{epsfig}
\usepackage{graphics}
\usepackage{algpseudocode}
\usepackage{amsmath}
\usepackage{pifont}
\newcommand{\tabincell}[2]{\begin{tabular}{@{}#1@{}}#2\end{tabular}}
\usepackage{algorithm}
\usepackage{hyperref}
\usepackage{exscale}
\usepackage{relsize}
\usepackage{setspace}

\usepackage[capitalize]{cleveref}

\crefname{section}{Sec.}{Secs.}
\Crefname{section}{Section}{Sections}
\Crefname{table}{Table}{Tables}
\crefname{table}{Tab.}{Tabs.}

\begin{document}

\title{Syntax-Aware Network for Handwritten Mathematical Expression Recognition}

\author{Ye Yuan$^{1*}$, Xiao Liu$^{1*}$, Wondimu Dikubab$^2$, Hui Liu$^1$, Zhilong Ji$^1$, Zhongqin Wu$^1$, Xiang Bai$^{2\ddagger}$\\
{\tt\small $^1$Tomorrow Advancing Life $^2$Huazhong University of Science and Technology}\\
{\tt\small \{yuanye8, jizhilong\}@tal.com, \{ender.liux, wondiyeaby, ryuki122382\}@gmail.com}\\
{\tt\small 30388514@qq.com,xbai@hust.edu.cn}}

\maketitle

\begin{abstract}
Handwritten mathematical expression recognition (HMER) is a challenging task that has many potential applications. Recent methods for HMER have achieved outstanding performance with an encoder-decoder architecture. However, these methods adhere to the paradigm that the prediction is made ``from one character to another'', which inevitably yields prediction errors due to the complicated structures of mathematical expressions or crabbed handwritings. In this paper, we propose a simple and efficient method for HMER, which is the first to incorporate syntax information into an encoder-decoder network. Specifically, we present a set of grammar rules for converting the LaTeX markup sequence of each expression into a parsing tree; then, we model the markup sequence prediction as a tree traverse process with a deep neural network. In this way, the proposed method can effectively describe the syntax context of expressions, 
alleviating the structure prediction errors of HMER. Experiments on two benchmark datasets demonstrate that our method achieves better recognition performance than prior arts. To further validate the effectiveness of our method, we create a large-scale dataset consisting of 100k handwritten mathematical expression images acquired from ten thousand writers. The source code$^{\S}$, new dataset$^{\dagger}$, and pre-trained models of this work will be publicly available.

\footnotetext[1]{Authors contribute equally.}
\footnotetext[3]{Corresponding author}
\footnotetext[4]{https://github.com/tal-tech/SAN}
\footnotetext[2]{https://ai.100tal.com/dataset}

\end{abstract}


\section{Introduction}

With the development of deep learning methods, the existing text recognition approaches are good at handling text lines in an image-to-sequence manner \cite{7801919, shi2018aster, shi2016robust, yang2019symmetry}. However, they may fail to deal with complicated structures such as mathematical expressions (ME). This paper investigates offline handwritten mathematical expression recognition (HMER), an important OCR task required by many applications like office automation, answer sheet correction, and assistance for visually disabled persons to understand mathematics. HMER is quite challenging, as a 2D structure relationship is essential for understanding mathematical expressions, which is seldom considered in previous deep learning-based methods. 
Besides, the ambiguities brought by handwritten input further increase the difficulty of HMER.

Early works have well studied ME's syntax structures, and the proper grammars are defined for HMER \cite{lavirotte1998mathematical, chan2001error, yamamoto2006line, alvaro2014recognition}. These grammars are only used for grouping the recognized symbols into a structural output, heavily relying on the performance of symbol recognition. Moreover, as these methods are mainly designed with handcraft features, their performance is far from the requirement of real-world applications.

\begin{figure}[h] 
    \centering
    \includegraphics[width=0.45\textwidth]{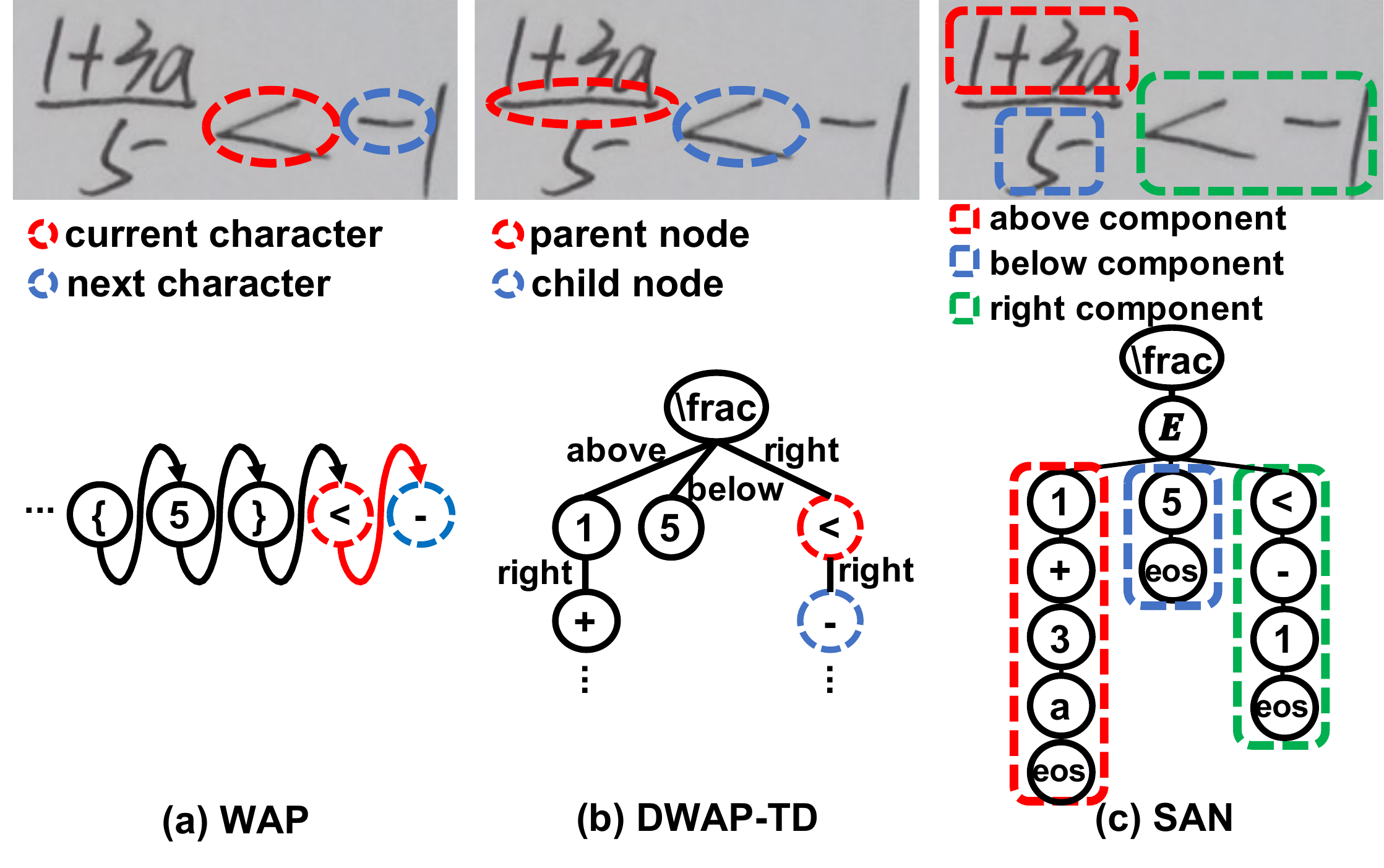}
    \caption{\textbf{Comparison of different architectures:} (a) An encoder-decoder framework WAP (b) A tree decoder DWAP-TD (c) Our model Syntax-Aware Network (SAN)}  
    \label{architectures comparison}
\end{figure}

Due to the recent advancement of deep neural networks, some recent studies \cite{deng2017image,zhang2017watch,zhang2018multi} handle HMER as an image-to-sequence prediction procedure using an encoder-decoder architecture, achieving significant performance improvements. However, these methods more or less neglect the syntax information contained in MEs.
To clearly illustrate this limitation, we take two recent network architectures as examples in Fig. \ref{architectures comparison}. Zhang \emph{et al.} \cite{zhang2017watch} propose the Watch, Attend and Parse (WAP) method that employs a fully convolutional network to encode handwritten images and a recurrent neural network as the string decoder to generate sequence outputs (Fig. \ref{architectures comparison}(a)). \cite{zhang2020tree} (DWAP-TD) attempt to consider the syntax information by decomposing the target syntactic structure tree into a sequence of sub-trees, where each sub-tree is composed of a parent node and a child node (Fig. \ref{architectures comparison}(b)). Though DWAP-TD can produce the output of a tree structure, it still follows the ``from one character to another'' paradigm that the next symbol prediction is mainly based on the current symbol. We argue that such methods don't explicitly consider the syntactic relationship of MEs in the learning process, which lacks the syntax constraints for generating a reasonable tree prediction.

To solve structure prediction error and improve the complex syntax tree understanding, we propose an elaborate grammar, which can naturally divide a syntax tree into different components and efficiently reduce tree structure ambiguity. Then, we establish an encoder-decoder network named syntax-aware network (SAN), which incorporates grammatical constraints and feature learning in a unified framework. Our intuition is that an ideal HMER model should parse handwritten mathematical expression images according to syntactic relationships, meanwhile effectively alleviating prediction errors caused by complex structures and crabbed writings. As shown in Fig. \ref{architectures comparison} (c), the prediction process of SAN follows the traverse process of a grammar tree, whose subtree is a significant component of a mathematical expression. In this manner, the syntactic relationship of adjacent components can be encoded in the proposed SAN model. Consequently, the prediction of SAN is made from one component to another component during the parsing procedure.

To evaluate the proposed SAN, we conduct the experiments on the two popular datasets, CROHME 2014\cite{mouchere2014icfhr} and CROHME 2016\cite{mouchere2016icfhr2016}. To further confirm the effectiveness of SAN, we collect and annotate a large-scale dataset for the evaluation, termed as HME100K. HME100K contains 100k handwritten mathematical expression images from ten thousand writers, mainly captured by cameras. Compared with the CHROME datasets\cite{mouchere2014icfhr,mouchere2016icfhr2016}, the data size of HME100K is increased tenfold. The results on CHROME 2014, CHROME 2016, and HME100K show that our method consistently achieves higher recognition rates over the state-of-the-art methods, demonstrating the advantage of  embedding syntax cues for HMER.

The main contribution of this paper is the proposed syntax-aware network, which effectively embeds syntactic information into deep neural networks at the first time. 
Another contribution of this paper is the proposed large and diverse dataset HME100K. Compared with the existing benchmark datasets, our dataset includes the HME with longer lengths and more complicated structures, making it useful to promote more robust algorithms toward real-world applications.

\begin{figure}[tp] 
    \centering
    \includegraphics[width=0.45\textwidth]{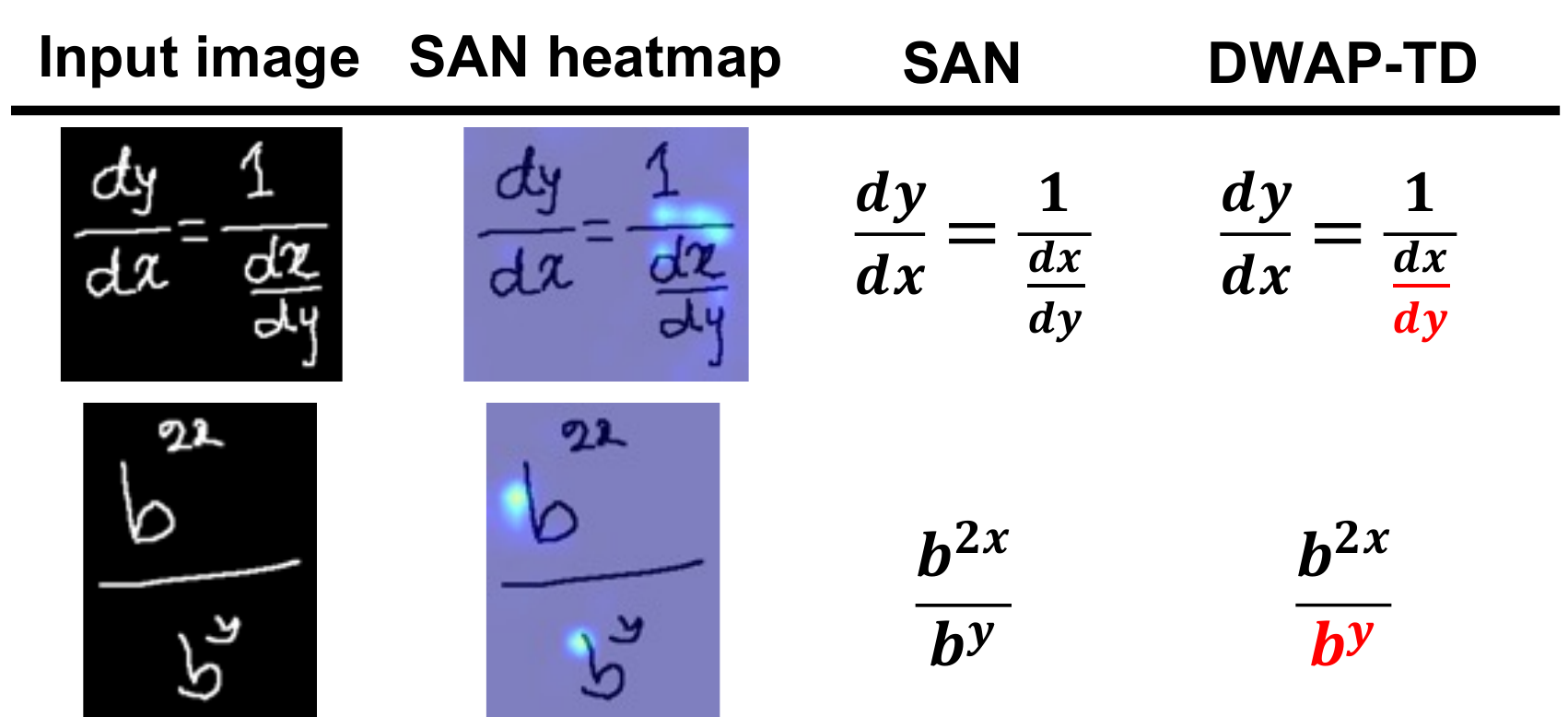}
    \caption{Sample Recognition Results of SAN and DWAP-TD. The SAN heatmaps indicate the model is focused on different components. The characters in red refer to the lost component during the prediction.}
    \label{bad cases}
\end{figure}

\section{Related Work}
HMER is a critical branches of document analysis and recognition that can be applied to recognize answer sheets of mathematics and digitize numerous kinds of scientific literature.
Compared with conventional handwriting text recognition, HMER is a more challenging task due to ambiguities coming from handwriting style, the two-dimensional structures complexity, and irregular scales of handwritten mathematics symbols. 
Therefore, HMER can be divided into three major challenging tasks\cite{chan2000mathematical,mouchere2016advancing, zanibbi2012recognition}: grouping strokes of the same symbol by segmentation, recognizing the symbols, and grammar-guided symbols structural relationship analysis to generate a mathematical expression. Traditional HMER methods tried to solve these challenges sequentially and globally.  


The sequential methods\cite{chan2000mathematical,zanibbi2002recognizing,winkler1996hmm, kosmala1999line, hu2011hmm, alvaro2014recognition,chan1998elastic, vuong2010towards,keshari2007hybrid } first segment input expression into mathematical symbols, classify each symbol separately, and then the structural relationship analysis recognize the mathematical expression. These methods employed classification techniques such as HMM \cite{winkler1996hmm, kosmala1999line, hu2011hmm, alvaro2014recognition}, Elastic Matching \cite{chan1998elastic, vuong2010towards}, Support Vector Machines \cite{keshari2007hybrid}, and tree transformations\cite{zanibbi2002recognizing}. 
On the other hand, the global approaches\cite{awal2014global, le2016system,alvaro2016integrated} apply a comprehensive strategy to learn mathematical symbols and their structural relationship analysis, while segmenting the symbols implicitly.
These methods handle HMER as a global optimization of mathematical expression segmentation, symbol recognition, and structure of the expression identification based on the symbol recognition results.

\begin{figure*}[h] 
\centering
\includegraphics[width=0.95\textwidth]{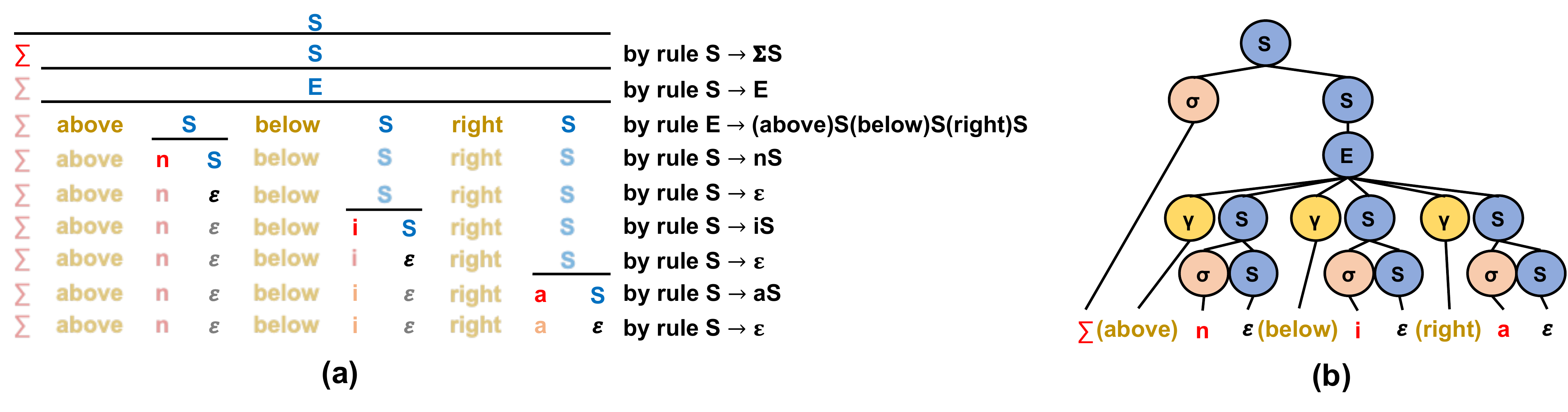}
\caption{(a) A possible parsing procedure of $\sum\limits_ {i}^{n}a$ and (b) the parse tree. In the figure, the strings refer to non-terminal symbols in blue, terminal symbols in red, relations in yellow, and empty in grey.}
\label{parsing procedure}
\end{figure*}

Recent deep learning-based HMER methods can be roughly classified into the sequence-to-sequence approach and the tree-structure approach. Most HMER methods extensively adopt the sequence-to-sequence approach.
The authors in\cite{deng2017image, zhang2018multi,zhang2017watch, zhang2018track, zhao2021handwritten, wang2019multi, truong2020improvement, nguyen2021temporal, le2020recognizing, zhang2018track, le2019pattern}  proposed an attention-based sequence-to-sequence model to convert the handwritten mathematical expression images into representational markup language LaTeX. Recently, Wu \emph{et al.}\cite{wu2021graph} designed a graph-to-graph(G2G) model that explores the HMEs structural relationship of the input formula and output markup, which significantly improve the performance.
The sequence-to-sequence approach improved the HMER performance. However, it lacks syntactic relationship awareness due to its inherited 1D nature, which causes inevitable prediction errors when dealing with 2D HMEs. 

The tree-structure approach employs a tree-structured decoder to model the parent-child relationship of mathematical expression trees explicitly.
To recognize online HME, Zhang \emph{et al.} \cite{zhang2017tree} proposed tree-structured based on BLSTM, while Zhang \emph{et al.} \cite{zhang2020srd} proposed tree-structured based on sequential relation decoder (SRD).  
On the other hand, to recognize offline HEMs, Zhang \emph{et al.} \cite{zhang2020tree} proposed a tree-structured decoder that attempts to consider the grammatical information by decomposing the target syntactic structure tree into a sub-tree sequence, where each sub-tree has a parent-child relationship.
Generally, a tree-structured decoder exhibits better robustness compared to a string decoder. However, the existing tree-structured decoders have two limitations. 1) The lack of holistic representation of grammatical information in deep network feature learning. 
2) The methods theoretically attempt to consider the syntactic information; however, they couldn't escape from the vicious circle of the sequence-to-sequence string decoding scheme in practice.    

The difficulty of HMER mostly relies on considering syntactic relationship complexity rather than symbol recognition. Thus, we proposed a novel neural network architecture equipped with grammar rules that efficiently divides a syntax tree into different components to alleviate errors caused by tree structure ambiguity.
The proposed method learns the grammatical relationship between the structures based on the grammar rules and navigates the syntax tree to create components according to the structural relationship. The key differences between our work and the existing tree-structured approaches are that (1) SAN integrates syntax constraints representation into deep neural network feature learning.  (2) Our method is a holistic syntax-aware approach that accurately identifies and locates components of MEs with complex structural relationships. As shown in Fig. \ref{bad cases}, SAN locates and recognizes the given MEs correctly while DWAP-TD misses a component of MEs. (3) SAN minimizes high computational costs by parsing one component to another instead of the individual parent and child nodes.

\section{ Syntax-Aware Network}

Formally, a SAN can be defined as a 7-tuple $G=(N, \Sigma, R, S, \Gamma, C, D)$, consisting of a finite set of non-terminal symbols ($N$), a finite set of terminal symbols ($\Sigma$), a finite set of production rules ($R$), a start symbol ($S$), a finite set of relations ($\Gamma$), an encoder ($C$) and a decoder ($D$). We design the grammar rules with two constraints: 1) It follows the standard reading orders: left-to-right, top-to-down. 2) The spatial relations between adjacent symbols are used. For a pair of adjacent symbols of HME, there are nine possible relations in total (left, right, above, below, low left, low right, upper left, upper right, inside). Due to constraint 1), we have removed ``left'' and ``low left'', and the rest 7 relations have been sustained to deal with all the cases of MEs in our implementation. Even though a ME may correspond to different LaTeX sequences, the syntax trees generated by our grammar rules are the same due to the two constraints.

The SAN transforms an image into a parse tree, of which the leaf nodes are terminal symbols or relations, and others are non-terminal symbols. There are two non-terminal symbols $S$ and $E$, where $S$ is the start symbol and servers as the tree's root. The terminal set $\Sigma$ contains all symbols that might be used in a LaTeX expression sequence.

The $R$ production rule can be used to construct the parse tree. The production rules are with the form of $\alpha\rightarrow\beta$, where $\alpha\in N$, $\beta \in (\Gamma \cup N \cup \Sigma) ^ *$ and the asterisk represents the Kleene star operation. Hence, a parent node ($\alpha$) can be split into a list of children nodes ($\beta$) containing terminal symbols, non-terminal symbols and relations.

$R$ has two production rules.
The first rule is an $S$ produce 1) an arbitrary terminal symbol followed by an $S$ on the right, 2) an $E$ or, 3) an empty string indicated by $\epsilon$:
\begin{equation}
    S\rightarrow\sigma S | E | \epsilon,
\end{equation}
where $\sigma\in\Sigma$ is an arbitrary terminal symbol and ``$|$'' separates alternatives. 
The second rule is an $E$ produces a string for each type of relation and then concatenates them. 
The string can be the relation followed by an S, or an empty string:
\begin{equation}
    E\rightarrow [((\gamma_1) S|\epsilon),\dots,((\gamma_7) S|\epsilon)],
    \label{node E}
\end{equation}
where $\gamma_i \in \Gamma$ is the $i^{th}$ type of relation in $\Gamma$, and [$\cdot$] is the concatenation operator. 

Fig. \ref{parsing procedure} illustrates the possible parsing procedure of an expression with the production rules.
To understand those rules intuitively, regard $S$ as an expression and $E$ an extendable structure. Assume that an expression may contain multiple extendable structures, whereas each extendable structure can be extended to multiple expressions with spatial relations. Moreover, the production rule is associated with a probability conditioned on the input image and the context state of the parent node.
Specifically, the conditional probability is defined as
\begin{equation}
p(\alpha\rightarrow\beta|c(\alpha),X) = D_{\alpha\rightarrow\beta}(c(\alpha),E(X)),
\end{equation}
where $X$ is the input image, $E(X)$ is the output of encoder, $c(\alpha)$ is the context state of $\alpha$ (will be detailed in Sec. 3.2) and $D_{\alpha\rightarrow\beta}(\cdot)$ is the output of decoder that corresponds to the production rule.

As illustrated in Algorithm \ref{algorithm}, given the SAN parameters and an input image, a tree traverse is implemented with a stack. Specifically, the implemented stack can guarantee the training process according to the traversal order on a syntax tree. Similarly, the prediction process is also implemented by stacking step by step.
The encoder takes an input image and down-samples it. Then based on the grammar rules, identify an expression and its extendable structures; meanwhile, the decoder calculates and selects the production rule with the highest probability. Consequently, generate new expressions with extendable structures and update the parse tree of the image in the LaTeX sequence.  
Once the parse tree is found, the recognition result can be acquired by traversing the tree in preorder. 
The remaining parts of this section focus on encoder, decoder, attention mechanism, and parameter learning. 

\begin{algorithm} 
    \setstretch{0.8}
    \caption{Inference process of SAN} 
    \label{algorithm}
    {\bf Input:} SAN parameters; The input image\\
    {\bf Output:} The parse tree of the image\\
    Encode the image\\
    Push $S$ and its context state onto the top of the stack
    \begin{algorithmic}
        \While{the stack is not empty} 
        {\\
            {\bf 1. }Pop a non-terminal symbol or relation together with its context state from the top of the stack\\
            {\bf 2.} Use  the  decoder  to  calculate  the  conditional probabilities of the production rules\\
            {\bf 3.} Select the production rule with the highest conditional probability\\
            {\bf 4.} Push each newly produced non-terminal symbol or relation in the rule with the context state onto the top of the stack\\
            {\bf 5.} Update the parse tree with the selected production rule
        }
        \EndWhile
    \end{algorithmic}
{\bf return} the parse tree
\end{algorithm}


\subsection{Encoder}
We use the densely connected convolutional network (DenseNet) \cite{huang2017densely} as the encoder. The DenseNet is an FCN that connects all networks in a feed-forward style and reinforces feature propagation and reuse by guaranteeing maximum information flow between layers in the network. Consequently, the FCN can handle an image of arbitrary size, making it appropriate for HMER since the sizes of handwritten images are usually in random sizes.
 
As a result, the encoder takes a gray-scale image $X$, with size of $1\times H\times W$, where $H$ and $W$ are height and width respectively, and returns a $C\times H/\zeta \times W/\zeta$ matrix, where $C$ is the channel number and $\zeta$ is the down-sampling factor.
Then the encoding is represented as $E(X) = [e_1,\ldots,e_L]$, where $L = H/\zeta \times W/\zeta$ and $e_i\in \mathbb{R}^C$.
Each element of $E(X)$ is related to a local region of the image.  
In our implementation, $C$ is set as 684 and $\zeta$ is set as 16.

\begin{figure}[h] 
\centering
\includegraphics[width=0.42\textwidth]{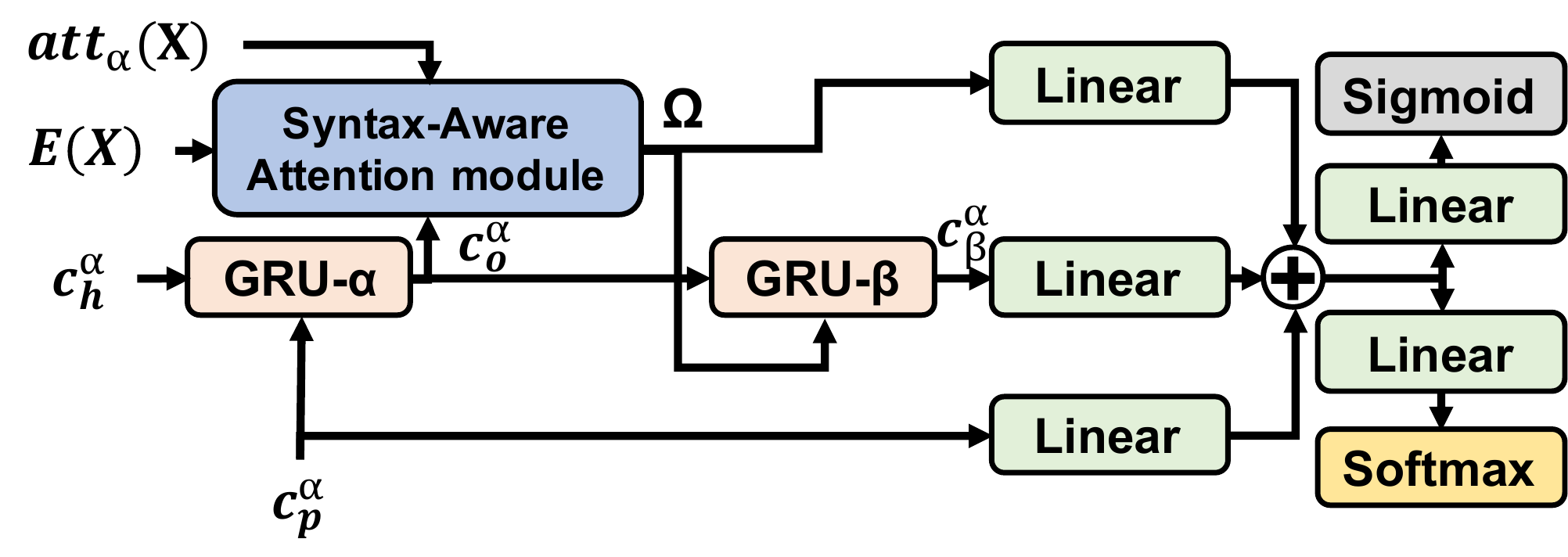}
\caption{Syntax-Aware Decoder: Consisting of   GRU-$\alpha$, GRU-$\beta$, and the Syntax-Aware Attention Module. }
\label{decoder}
\end{figure}
\vspace{-0.3cm}

\subsection{Syntax-Aware Decoder}
As illustrated in Fig. \ref{decoder}, the decoder mainly consists of two Gated Recurrent Units (GRU) cells \cite{chung2014empirical} (indicated by GRU-$\alpha$, GRU-$\beta$) and a syntax-aware attention module.
Given the context state of a non-terminal symbol $\alpha$ and the encoding vector $E(X)$, the decoder returns the probability of each production rule that begins with $\alpha$.

We use a historical state and a partner state to describe the context state of the current parsing non-terminal symbol. 
The historical state $c_h^{\alpha}$ is used to keep track of the information of how the non-terminal symbol $\alpha$ is produced.
In addition, the word embedding of the latest generated terminal symbol or relation is used as the partner state $c_p^{\alpha}$ of the non-terminal symbol to capture the short-term contextual information. 
The first GRU-$\alpha$ takes $c_p^{\alpha}$ as the input vector and  $c_h^{\alpha}$ as the hidden vector, and outputs a new hidden vector $c_o^{\alpha}$:
\begin{equation}
    c_{o}^{\alpha} = \mathrm{GRU}(c_{p}^{\alpha}, c_{h}^{\alpha}).
\end{equation}
Then the attention module calculates a compact visual feature
\begin{equation}
\Omega =  \mathrm{Att}(E(X),c_{o}^{\alpha},att_{\alpha}(X)),
\end{equation}
where $\mathrm{Att}(\cdot)$ is the attention function and 
$att_{\alpha}(X)$ is the syntax-aware attention vector
which is detailed in Sec 3.3. 
The second GRU-$\beta$ takes $\Omega$ as the input vector and $c_o^{\alpha}$ as the hidden vector, and outputs a new hidden vector $c_\beta^{\alpha}$:
\begin{equation}
    c_\beta^{\alpha} = \mathrm{GRU}(\Omega,c_{o}^{\alpha}).
\end{equation}
We then aggregate $c_p^{\alpha}$, $c_\beta^{\alpha}$ and $\Omega$ to predict two branches of probabilities:
\begin{equation}
    \begin{split}
        p_{symbol}(\alpha\rightarrow\beta|c(\alpha),X) = \hspace{0.9in}\\
        \mathrm{softmax}(W_{s}(W_{p}c_p^{\alpha}
        + W_{g}c_\beta^{\alpha} + W_{t}\Omega))
    \end{split}
\end{equation}
\vspace{-0.5cm}
\begin{equation}
    \begin{split}
        p_{relation}(\alpha\rightarrow\beta|c(\alpha),X) = \hspace{0.9in} \\ \mathrm{sigmoid}(W_{r}(W_{p}c_p^{\alpha}
        + W_{g}c_\beta^{\alpha} + W_{t}\Omega))
    \end{split}
\end{equation}
where $W_s$, $W_p$, $W_g$, $W_t$ and $W_r$ are learnable parameters. 
$p_{symbol}(\alpha\rightarrow\beta|c(\alpha),X)$ is a probability vector with $|\Sigma| + 2$ dimensions. Note that there are three scenarios 1) $|\Sigma|$ dimensions for predicting terminal symbols, 2) one dimension for predicting $E$, and 3) one dimension for predicting an empty string.
For the first scenario, if the prediction of a terminal symbol ($\sigma$) has the highest probability, then apply the rule of $S\rightarrow \sigma S$ to update the parse tree, use the word embedding of $\sigma$ as the partner state of the newly generated $S$, and $c_\beta^{\alpha}$ as its historical state.
For the second scenario, if the prediction of $E$ has the highest probability, then apply $p_{relation}(\alpha\rightarrow\beta|c(\alpha),X)$ to predict the probability of each relation.
The relations with probabilities higher than 0.5 are sustained, and others are thrown away. 
For each remaining relation, we use the embedding of the relationship as the partner state of the following $S$, and use $c_\beta^{\alpha}$ as the historical state. Moreover, there is no need to consider parsing an $E$, because it has already been acquired from the relation branch.
For the third scenario, 
if the empty string has the highest probability, we update the parse tree with the rule of $S\rightarrow \epsilon$.

\subsection{Syntax-Aware Attention Module}
Instead of using the entire image feature for decoding, the attention module calculates a compact visual feature with the attention mechanism. 
We first compute a normalized weight for each local region of the image, and then use the weighted average to aggregate the local features.
We use the image encoding $E(X)$, the hidden state $c_o^{\alpha}$ and a syntax-aware attention vector $att_{\alpha}(X)$ to calculate the normalized weight vector:
\begin{equation}
\begin{split}
        \xi_\alpha = \mathrm{softmax}(W_{w}( \mathrm{tanh}(W_{o}c_{o}^{\alpha} +  \\ W_{\alpha}att_{\alpha}(X) + W_{e}E(x)))),
\end{split}
\end{equation}
where $W_w$, $W_o$, $W_{\alpha}$ and $W_e$ are learnable parameters, and $\xi_\alpha$ is a vector of length $L$. 
The compact visual feature 
\begin{equation}
\mathrm{Att}(E(X),c_{o}^{\alpha},att_{\alpha}(X)) = E(X)\xi_\alpha
\end{equation}
is computed by the matrix product. 
Unlike \cite{zhang2017watch,zhang2020tree}, which used a coverage vector based on the sum of all past attention probabilities, SAN does not keep track of all past attention probabilities.
The attention drift problem happens because the attention probabilities of the numerator give no information for parsing the denominator but appear as the noise.  
Instead, we sum up the past attention probabilities along the path from the root of the parse tree to the current parsing node but not all past attention probabilities.
Therefore, we calculate the syntax-aware attention vector as follows:
\begin{equation}
att_{\alpha}(X) = \sum_i \xi_i, \hspace{0.1in} i\in path_{\alpha}.
\end{equation}

The syntax-aware attention vector can be efficiently traced by storing it as intermediate information with the stack. 

An attention self-regularization strategy is used to correct the attention. We use an additional reversed decoder to predict the parent node of each given child node.
The reversed decoder shares the same structure with the original one but operates the data in reverse. 
Thus we have two normalized weight vectors for predicting each non-terminal node $\beta$, the forward $\xi_\alpha$ and the reversed $\hat{\xi_\eta}$, where $\alpha$ is the parent of $\beta$ and $\beta$ is the parent of $\eta$.
We use a Kullback-Leibler (KL) divergence to regularize them
\begin{equation}
     \mathcal{L}_{reg} = -\sum \limits_\beta  \hat{\xi_\eta} \log \frac{\hat{\xi_\eta}}{\xi_\alpha}.
\end{equation}
The reversed decoder is jointly trained with SAN but is omitted during inference to skip additional inference time.

\begin{table*}[htp]
    \renewcommand\arraystretch{0}
    \centering
    \caption{Statistical comparison of the HME100K and CROHME2019 dataset. ``Max Length'' and ``Avg Length'' mean the maximum length and average length of the mathematical expression.}
    \label{statistics}
    \begin{tabular}{c|cccccc}
    \toprule[1.5pt]
    \textbf{Dataset}             & \textbf{Train Size}& \textbf{Test Size}&  \textbf{Symbol Class No.} & \textbf{Max Length} & \textbf{Avg Length} & \textbf{Writer No.} \\ \midrule[1pt]
    HME100K     & 74,502 & 24,607& 245 & 184 & 17.62 & $\thicksim$10K \\ \midrule[1pt]
    CROHME2019 & 8,836 & 1,199& 101 & 96 & 15.79 & $\thicksim$100 \\
    \bottomrule[1.5pt]
    \end{tabular}
\end{table*}

\subsection{Parameter Learning}
SAN is trained end-to-end under a multi-task setting, that aims to minimize the sum of symbol loss ($\mathcal{L}_{symbol}$), relation loss ($\mathcal{L}_{relation}$), reversed symbol loss ($\mathcal{L}_{symbol}^{rev}$) and attention self-regularization loss ($\mathcal{L}_{reg}$).

We use the teacher forcing strategy for accelerating convergence. The ground-truth parse tree is obtained for each training image by parsing the LaTeX sequence with the depth-first-search algorithm.  Thus, a list of parent-children samples is acquired from the parse tree. Afterward, deal with each sample sequentially according to the preorder of the tree until the entire tree is processed. Likewise, we use the reversed decoder to predict the parent node of each given child node. 
Then using the ground-truth labels, we calculate the symbol, reversed symbol and the relation losses.
The attention self-regularization loss can be calculated by (12). Thus we can optimize the parameters by minimizing the following objective function with backpropgation
\begin{equation}
    \mathcal{L} = \mathcal{L}_{symbol} + \mathcal{L}_{relation} + \mathcal{L}_{symbol}^{rev} + \mathcal{L}_{reg}.
\end{equation}

\section{The HME100K Dataset}
This section presents the new dataset HME100K, consisting of 74,502 images for training and 24,607 images for testing with 245 symbol classes, as shown in Table \ref{statistics}. Compared to CRHOME 2019\cite{mahdavi2019icdar} dataset, the data size is increased tenfold. The data were collected from tens of thousands of writers who wrote the MEs on papers and uploaded them to an internet application.   

\begin{figure}[h] 
    \centering
    \includegraphics[width=0.43\textwidth]{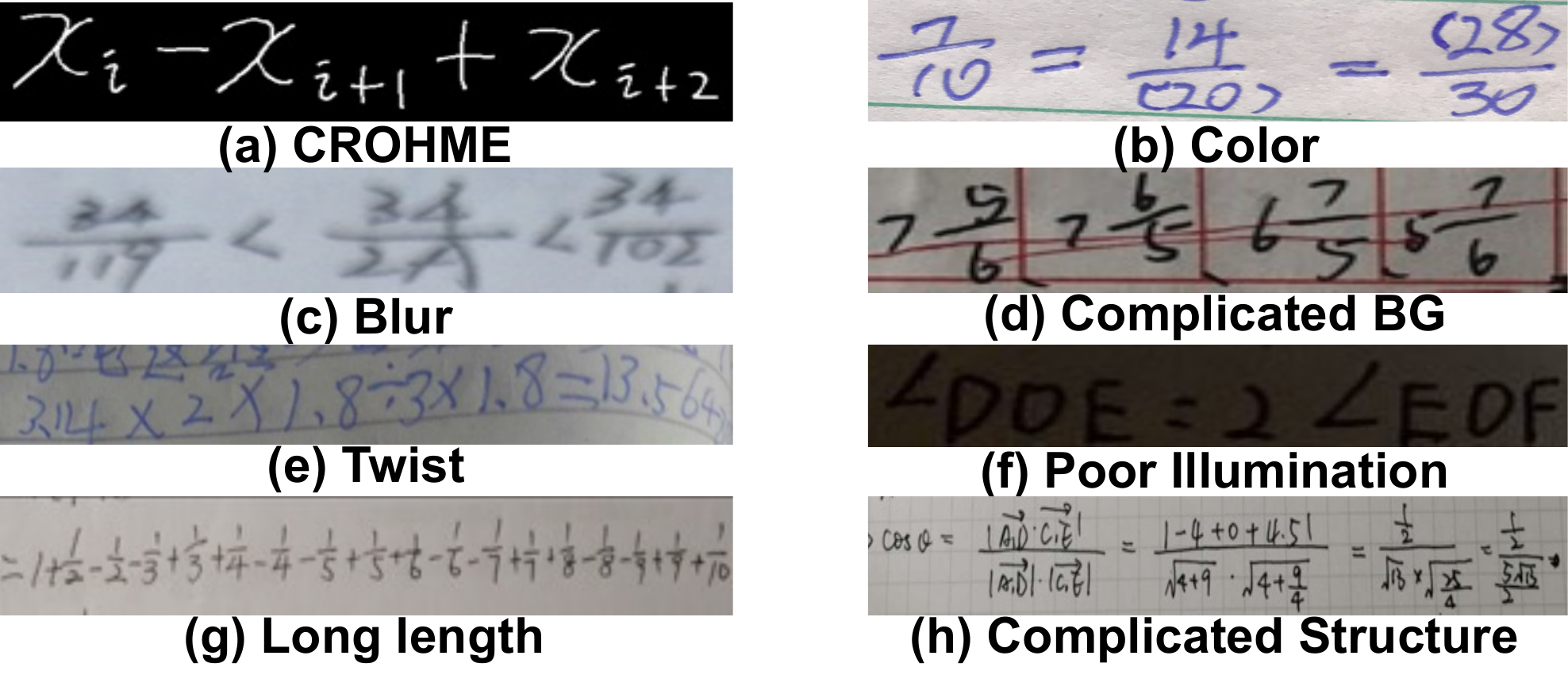}
    \caption{Sample images from (a) the CROHME dataset and (b-h) the HME100K dataset.} 
    \label{dataset}
\end{figure}

As tens of thousands of writers participate in writing MEs, the variety of writing styles creates unique features to our MEs dataset.  Consequently, our dataset HME100K becomes more authentic and realistic with variations in color, blur, complicated background, twist, illumination, longer length, and complicated structure. Samples images from the dataset are shown in Fig. \ref{dataset}(b-h).
Furthermore, the maximum character length in HME100K is 184, which is almost twice longer than CROHME. HME100K also has a longer average sequence length than CROHME. For additional information about the HME100K dataset, please refer to the supplementary materials.



\section{Experiments}
We evaluate our method performance on two CROHME benchmark datasets and the proposed HME100K and make the comparison with the state-of-the-art methods. 

\begin{table*}[htbp]
    \renewcommand\arraystretch{0.3}
    \centering
    \caption{Expression Recognition Rate (ExpRate) and Expression Structure Prediction Rate (ESPR) performance of our model and other state-of-the-art methods on CROHME 2014 and CROHME 2016. All results are reported as a percentage (\%). Our model achieves the best performance on all public datasets. $\star$ indicates the methods that used data augmentation during the training process.}
    \label{contrast results}
    \begin{tabular}{c|ccc|ccc|ccc}
    \toprule[1.5pt]
    \multirow{2}{*}{\textbf{Method}} & \multicolumn{3}{c|}{\textbf{CROHME 2014}} & \multicolumn{3}{c|}{\textbf{CROHME 2016}} & \multicolumn{3}{c}{\textbf{CROHME 2019}} \\
    
     & \textbf{ExpRate}&\textbf{$\leq$1}&\textbf{$\leq$2}&\textbf{ExpRate}&\textbf{$\leq$1}&\textbf{$\leq$2}&\textbf{ExpRate}&\textbf{$\leq$1}&\textbf{$\leq$2}\\ \midrule[1pt]
    
    WYGIWYS \cite{deng2017image}&       36.4& -& -& -& -& -& -& -& -\\
    WAP \cite{zhang2017watch}&          40.4& 56.1& 59.9& 44.6& 57.1& 61.6& -& -& -\\
    PAL \cite{wu2018image}&             39.7& 56.8& 65.1& -& -& -& -& -& -\\
    PAL-v2 \cite{wu2020handwritten}&    48.9& 64.5& 69.8& 49.6& 64.1& 70.3& -& -& -\\
    PGS \cite{le2019pattern} & 48.8& 66.1& 73.9& 36.3& -& -& -& -& -\\
    TAP \cite{zhang2018track}& 48.5& 63.3& 67.3& 44.8& 59.7& 62.8& -& -& -\\
    DLA \cite{le2020recognizing}& 49.9& -& -& 47.3& -& -& -& -& -\\
    DWAP \cite{zhang2018multi}&     50.1& -& -& 47.5& -& -& -& -& -\\
    DWAP-MSA \cite{zhang2018multi}& 52.8& 68.1& 72.0& 50.1& 63.8& 67.4& 47.7& 59.5& 63.3\\
    DWAP-TD \cite{zhang2020tree}&   49.1& 64.2& 67.8& 48.5& 62.3& 65.3& 51.4& 66.1& 69.1\\
    WS-WAP \cite{truong2020improvement}& 53.7& -& -& 52.0& 64.3& 70.1& -& -& -\\
    MAN \cite{wang2019multi}& 54.1& 68.8& 72.2& 50.6& 64.8& 67.1& -& -& -\\
    RBR \cite{truong2020improvement}& 53.4& 65.2& 70.3& 52.1& 63.2& 69.4& 53.1& 63.9& 68.5 \\
    DWAP + CTC \cite{nguyen2021temporal}& 51.0& -& -& 50.0& -& -& -& -& -\\
    BTTR \cite{zhao2021handwritten}& 54.0& 66.0& 70.3& 52.3& 63.9& 68.6& 53.0& 66.0& 69.1 \\
    Li \emph{et al.} \cite{DBLP:conf/icfhr/LiJLZ20}$^\star$ & 56.6& 69.1& 75.3& 54.6& 69.3& 73.8& -& -& - \\
    Ding \emph{et al.} \cite{10.1007/978-3-030-86331-9_39}$^\star$ & 58.7& -& -& 57.7& 70.0& 76.4& 61.4& 75.2& 80.2 \\
    \textbf{SAN}&  \textbf{56.2}& \textbf{72.6}& \textbf{79.2}& \textbf{53.6}& \textbf{69.6}& \textbf{76.8}& \textbf{53.5}& \textbf{69.3}& \textbf{70.1}\\ 
    \textbf{SAN}$^\star$&  \textbf{63.1}& \textbf{75.8}& \textbf{82.0}& \textbf{61.5}& \textbf{73.3}& \textbf{81.4}& \textbf{62.1}& \textbf{74.5}& \textbf{81.0}\\ 
    \bottomrule[1.5pt]
    \end{tabular}
\end{table*}

\subsection{Datasets}
We use the Competition on Recognition of Online Handwritten Mathematical
Expressions (CROHME) benchmark, which currently is the most widely used public dataset for
handwritten mathematical expression to train and validate the HMER model. 
We convert the handwritten stroke trajectory information in the InkML files into image format for the training and test sets. Sample of CROHME images is shown in (Fig. \ref{dataset}(a)). 

The CROHME training set consists of 8,836 expressions, including 101 math symbol classes, while the test set number of expressions are different according to the year they were published. 
We evaluate our model on CROHME 2014 test set\cite{mouchere2014icfhr}, CROHME 2016 test set \cite{mouchere2016icfhr2016} and CROHME 2019 test set \cite{mahdavi2019icdar} contain 986, 1147 and 1199 expressions respectively.


\subsection{Implementation Details}
The proposed model SAN is implemented in PyTorch.  All experiments are conducted on a single Nvidia Tesla V100 with 32GB RAM, and the batch size is set at 8.
The hidden state sizes of the two GRUs are set at 256. The dimensions of word embedding and relation embedding are set at 256. 
The Adadelta optimizer \cite{zeiler2012adadelta} is used during the training process, in which $\rho$ is set at 0.95 and $\epsilon$ is set at $10^{-6}$. 
The learning rate starts from 0 and monotonously increases to 1 at the end of the first epoch. 
After that the learning rate decays to 0 following the cosine schedules \cite{zhang2019bag}. 
Following most previous works, no data augmentation is applied during training for the fair comparisons. 

\subsection{Evaluation Protocol}
\noindent\textbf{Recognition Protocol.} Expression recognition rate (ExpRate) is the widely used recognition protocol for mathematical expression recognition, defined as the percentage of predicted mathematical expressions accurately matching the ground truth. ExpRate $\leq$ 1 and $\leq$ 2 indicate the expression recognition rates are tolerable at most one or two symbol-level errors.

\noindent\textbf{Structure Recognition Protocol.} Expression Structure Prediction Rate (ESPR) is used as the structure recognition protocol. ESPR is calculated by the percent of MEs whose structure is recognized correctly irrespective of symbol labels.

\subsection{Comparisons with State-of-the-arts}
In this subsection, we measure our proposed method on CROHME 2014, CROHME 2016 and CROHME 2019 datasets and compare the performance with other state-of-the-art methods. Most of the previous methods do not use data augmentation, so we mainly focus on the results produced without data
augmentation. Note that our method is not with the beam search process to obtain additional performance improvements.

\noindent\textbf{Evaluation on CROHME dataset}
As shown in table \ref{contrast results}, our method achieves state-of-the-art performance on all CROHME datasets. SAN outperforms MAN \cite{wang2019multi} by 3.1\%, BTTR \cite{zhao2021handwritten} by 1.3\% and BTTR \cite{zhao2021handwritten} by 0.5\% on CRHOME 2014, 2016 and 2019, respectively. In addition to ExpRate, SAN outperforms other state-of-the-art methods by a larger margin in ESPR. Thus, the SAN achievement demonstrates that incorporating syntax information into HMER neural network is effective and efficient.

\begin{table}[h]
    \renewcommand\arraystretch{0.1}
    \setlength\tabcolsep{2pt}
    \centering
    \caption{Performance of our model versus DWAP, DWAP-TD and BTTR on Easy (E.), Moderate (M.) and Hard (H.) HME100K test subsets. Inference speed reported as FPS means frames per second. The last column shows the parameter numbers of each model. The number in bold font corresponds to the best performance, and the second-best result is shown with an underline. Our model achieves the best performance on the HME100K dataset.}
    \label{subset performance}
    \begin{tabular}{c|cccccc}
    \toprule[1.5pt]
    \textbf{HME100K}& \textbf{Easy}& \textbf{Moderate}& \textbf{Hard}& \textbf{Total}& \textbf{FPS}& \textbf{P.N.} \\ \midrule[1pt]
    Image size&       7721&   10450&   6436& 24607& -&-\\\midrule[0.5pt]
    DWAP\cite{zhang2018multi}&    75.1&  62.2& 45.4& 61.9& \underline{23.3}& 4.8M\\
    DWAP-TD\cite{zhang2020tree}& 76.2& 63.2& 45.4& 62.6& 6.9 & 8.0M\\
    BTTR \cite{zhao2021handwritten}& \underline{77.6} & \underline{65.3} & \underline{46.0} & \underline{64.1} & 3.9 & 6.5M\\
    SAN&                      \textbf{79.2}&  \textbf{67.6}& \textbf{51.5}& \textbf{67.1}& \textbf{23.9} & 8.9M\\
    \bottomrule[1.5pt]
    \end{tabular}
\end{table}

\subsection{Evaluation on HME100K}

\noindent \textbf{Dataset Division.} For mathematical expressions, the structural complexity (S.C.) \cite{zhang2017tree} and character length (C.L.) significantly affect the model performance. With this in mind, we carefully allocate our test dataset into three subsets by difficulty. The allocation criteria is as follow:

$\begin{cases}
\textbf{Easy}, \qquad S.C. \in [0, 1] \ \& \ C.L. \in [1, 10) \\
\textbf{Moderate}, \ \ S.C. \in [0, 1] \ \& \ C.L. \in [10, 20) \\
\textbf{Hard}, \qquad otherwise
\end{cases}$

\noindent\textbf{Comparisons with previous methods.} In this subsection, we compare our proposed method with DWAP \cite{zhang2018multi}, DWAP-TD \cite{zhang2020tree} and BTTR \cite{zhao2021handwritten} on HME100K dataset. In order to be consistent with the reported inference process, DWAP-TD \cite{zhang2020tree} and BTTR \cite{zhao2021handwritten} use beam search, while DWAP \cite{zhang2018multi} doesn't. Specifically, as shown in table \ref{subset performance}, our method outperforms BTTR \cite{zhao2021handwritten} by 1.6\% on easy subset. However, as the difficulty of the test subset increases, the leading margin of our method increases to 5.5\% on the hard subset. The measurements of performances on different test subsets prove the superior robustness and structural awareness of our method.

In addition to the recognition accuracy, we evaluate the inference speed of our model against DWAP \cite{zhang2018multi}  and DWAP-TD \cite{zhang2020tree}  as summarised in table \ref{subset performance}. We measure the inference speeds in frame per second (FPS) on an HME100K test set via an Nvidia Tesla V100. Impressively, SAN is 3.5 times faster than DWAP-TD \cite{zhang2020tree}, while 2.6\% faster than DWAP \cite{zhang2018multi}, which indicates the efficiency of our method in minimizing computational costs.

\begin{table}[h]
    \renewcommand\arraystretch{0.5}
    \centering
    \caption{Ablation studies on CROHME datasets and HME100K datasets. The effect of recognition performance with regard to the two basic parameter settings: grammar syntax (G.S.) and syntax-aware attention (S.A. Attention). (\ding{51}) denotes the module existence while (\ding{53}) indicates the module absence.}
    \label{ablation studies}
    \begin{tabular}{c|ccccc}
    \toprule[1.5pt]
    \textbf{Model} & \textbf{G.S.} & \textbf{S.A.} & \textbf{ExpRate} & \textbf{Dataset} \\ \midrule[1pt]
    Baseline & \ding{53}&  \ding{53}& 49.1 &\multirow{3}{*} {\tabincell{c}{CROHME\\2014}}\\
    SAN\_GS  & \ding{51}&  \ding{53}& 55.3\\  
    SAN      & \ding{51}&  \ding{51}& 56.2\\ \midrule[1pt]
    
    Baseline & \ding{53}&  \ding{53}& 48.5 &\multirow{3}{*} {\tabincell{c}{CROHME\\2016}}\\
    SAN\_GS  & \ding{51}&  \ding{53}& 52.8\\  
    SAN      & \ding{51}&  \ding{51}& 53.6\\ \midrule[1pt]
    
    Baseline & \ding{53}&  \ding{53}& 51.4 &\multirow{3}{*} {\tabincell{c}{CROHME\\2019}}\\
    SAN\_GS  & \ding{51}&  \ding{53}& 53.0\\  
    SAN      & \ding{51}&  \ding{51}& 53.5\\ \midrule[1pt]
    
    Baseline & \ding{53}&  \ding{53}& 62.6 &\multirow{3}{*} {HME100K}\\
    SAN\_GS  & \ding{51}&  \ding{53}& 66.5\\  
    SAN      & \ding{51}&  \ding{51}& 67.1\\
    \bottomrule[1.5pt]
    \end{tabular}
\end{table}

\subsection{Ablation Study}
In this subsection, we perform ablation studies to analyze the impact of grammar syntax and syntax-aware attention modules. We trained all models from scratch and evaluated their performance on three datasets. SAN is the default model, while SAN-GS has the syntax-aware decoder but adopts coverage attention in the attention module instead of syntax-aware attention. The results are summarized in Table \ref{ablation studies}.

\noindent \textbf{Impact of Grammar Syntaxes.} 
Table \ref{ablation studies} shows the integration of grammar syntaxes to baseline improves the recognition performance ExpRate by 6.2\% on CROHME 2014, 4.3\% on CROHME 2016, and 3.9\% on HME100K. Hence integrating grammatical constraints into the baseline model achieved improvements on all datasets, and this proofs the significance of grammar syntaxes incorporation to the decoder.   

\begin{figure}[h] 
    \centering
    \includegraphics[width=0.4\textwidth]{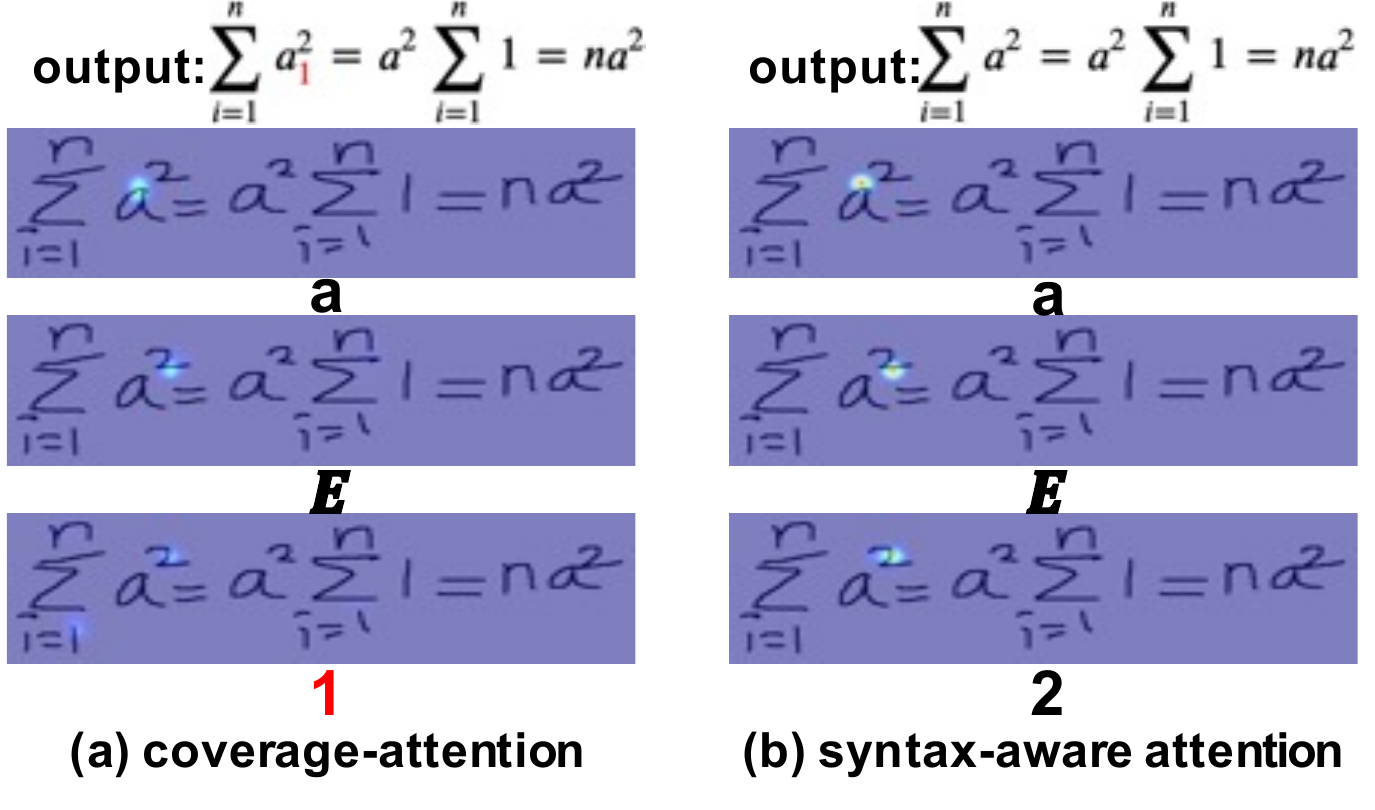}
    \caption{Examples of (a) Converge-Attention and (b) Syntax-Aware Attention. The current recognized symbol is printed below each image. ``$E$'' indicates the non-terminal symbol (Eq.(1)). } 
    \label{grammar aware attention}
\end{figure}

\noindent \textbf{Impact of Syntax-Aware Attention.} As it is illustrated in Table \ref{ablation studies}, incorporating syntax-aware attention on top of grammar syntaxes increases the recognition performance rate ExpRate by 7.1\% on CROHME 2014, 5.1\% on CROHME 2016, and 4.5\% on HME100K from the baseline.

%
Furthermore, as shown in Fig. \ref{grammar aware attention}, we compare the coverage attention and syntax-aware attention through a qualitative example. The images extracted from the prediction steps of coverage-attention and syntax-aware attention models illustrate how each model focuses on the target at each step.  As shown in (Fig. \ref{grammar aware attention}(a)), the coverage attention model wrongly focuses on the region of the symbol ``1", predicting a redundant sub-tree that doesn't exist in the predicted position. 
In contrast, the syntax-aware attentions model predicts an appropriate location that coincides with human intuitions (Fig. \ref{grammar aware attention}(b)).


    

\section{Limitation}
\begin{figure}[h] 
    \centering
    \includegraphics[width=0.48\textwidth]{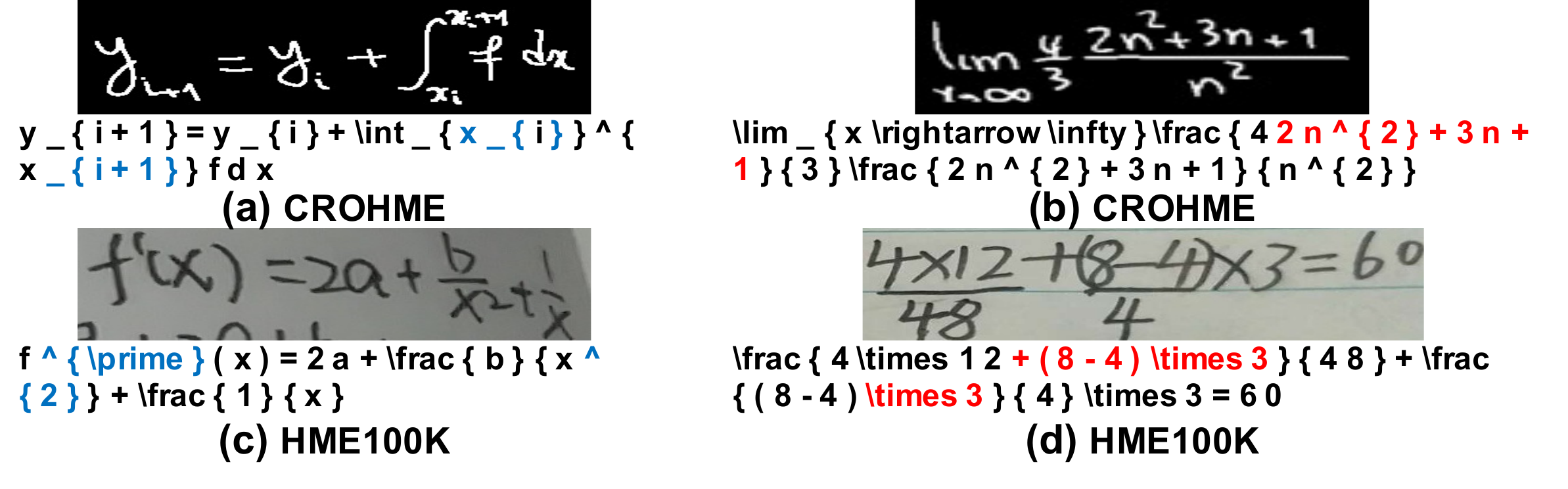}
    \caption{Limitation sample examples with the ground truth and recognition results of SAN. Characters in red color are mispredictions.} 
    \label{Histograms}
\end{figure}

As it is illustrated in Fig. \ref{Histograms}, distorted and sticking components of ME can cause failure to SAN prediction, which leads to the under/over translation. Fig. \ref{Histograms} gives four typical examples on CROHME (a and b) and HME100K (c and d) dataset.

\section{Conclusions}
This paper has presented an unconventional method for handwritten mathematical expression recognition by combining syntax information and visual representations to make robust predictions. To our best knowledge, the proposed syntax-aware network is the first to effectively incorporate the grammar rules into deep feature learning. Our method not only predicts LaTeX markup results but also directly produces the tree structure output that can precisely describe the component relationship of mathematical expressions. Experiments on the benchmark datasets and the proposed HME100K dataset have validated the effectiveness and efficiency of our method. In our future work, we are interested in extending the proposed method to recognize other complicated structure objects.

\section{Acknowledgement}

This work was supported by National Key R\&D Program of China, under Grant No. 2020AAA0104500 and National Natural Science Foundation of China 61733007.

{\small
\bibliographystyle{ieee_fullname}
\bibliography{egbib}
}

\end{document}